\documentclass[11pt]{article}

\usepackage[letterpaper,margin=1in]{geometry}
\usepackage{setspace}
\usepackage{times}
\usepackage{latexsym}
\usepackage[authoryear,round]{natbib}
\usepackage[hidelinks]{hyperref}

\usepackage[T1]{fontenc}

\usepackage[utf8]{inputenc}

\usepackage{microtype}

\usepackage{inconsolata}

\usepackage{graphicx}

\usepackage{algorithm}
\usepackage{algorithmic}
\usepackage{subcaption}
\usepackage{graphicx}
\usepackage{amsmath}
\usepackage{amssymb}
\usepackage{booktabs}
\usepackage{ragged2e}
\usepackage{array}
\usepackage{wrapfig}

\usepackage[hang,flushmargin]{footmisc}
\usepackage{newfloat}
\usepackage{listings}

\title{CVPO: Enhancing LLM Reinforcement Learning Reasoning via Value-Variance Adaptation and Dynamic Curriculum Learning}

\author{
 \textbf{Ziqi Jia\textsuperscript{1,2,$\blacklozenge$}},
 \textbf{Yalu Ouyang\textsuperscript{1,3,$\blacklozenge$}},
 \textbf{Bo Pang\textsuperscript{1}},
 \textbf{Panpan Li\textsuperscript{1}},
\\
 \textbf{Hangfei Xu\textsuperscript{1}},
 \textbf{Shengzhao Wen\textsuperscript{1}},
 \textbf{Shiyong Li\textsuperscript{1}},
 \textbf{Yanpeng Wang\textsuperscript{1}},
\\
 \textsuperscript{1}AI Cloud Group, Baidu,
 \textsuperscript{2}Tsinghua University,
 \textsuperscript{3}University of California, San Diego
\\
 \small{
   \textbf{Correspondence:} \href{mailto:jzq_2022@163.com}{jzq\_2022@163.com}
 }
}

\begin{document}
\date{}
\maketitle
\begin{abstract}
Reinforcement learning (RL) has emerged as an effective method for enhancing the reasoning capabilities of large language models (LLMs). However, existing methods suffer from insufficient precision in feedback on generated answer trajectories and exhibit the phenomenon of problem difficulty drift. To address these challenges, we propose \textbf{CVPO}\textemdash \textbf{C}urriculum-guided \textbf{V}alue-\textbf{V}ariance \textbf{P}olicy \textbf{O}ptimization. At the response trajectory level, we find that token-level value-variance correlates with exploration intensity. Our theoretical analysis shows this variance bounds policy update magnitude. We then use the estimated trajectory value-variance to quantify the intrinsic randomness in generation. Based on this, we design a variance-aware advantage adjustment mechanism for different reward types. At the question level, we introduce a dynamic curriculum weighting method that adapts to question difficulty. This helps the model focus on tasks matched to its current ability during each training stage. Experimental results show our method outperforms strong value-based baselines like VAPO. It achieves better performance and stronger exploration, enabling more accurate and robust reasoning in language models across various math tasks.
\end{abstract}

\section{Introduction}

In recent years, reasoning models based on reinforcement learning have demonstrated remarkable performance breakthroughs in solving high-difficulty mathematical reasoning tasks, providing a new approach for solving complex logical problems \cite{DBLP:conf/acl/AhmadianCGFKPUH24,DBLP:journals/corr/abs-2501-03262,DBLP:journals/corr/abs-2501-12599,DBLP:journals/corr/abs-2412-19437,DBLP:journals/corr/abs-2506-13585,DBLP:journals/corr/abs-2505-12951,DBLP:journals/corr/abs-2504-13914,DBLP:journals/corr/abs-2405-19107}. Among them, value-free model methods represented by GRPO \cite{2025deepseek-math} have achieved pioneering progress in related tasks by averaging the rewards of multiple trajectories within a group to calculate advantage values. However, subsequent studies on value-based algorithms such as VAPO \cite{DBLP:journals/corr/abs-2504-05118} have further revealed that well-trained value models can achieve more accurate credit assignment, and their performance upper limit is significantly higher than that of value-free model methods such as GRPO, pointing out a more promising direction for the development of this field. 

\begingroup
\renewcommand{\thefootnote}{\ensuremath{\blacklozenge}}%
\footnotetext[1]{Completed during internship at Baidu}%
\endgroup

Nevertheless, current value-based methods still have two key bottlenecks that needs addressing. First, the feedback accuracy on the generated answer trajectories is insufficient \cite{DBLP:journals/corr/abs-2505-22660,DBLP:journals/corr/abs-2506-10947}. When the trajectories show homogeneous results (all correct or all incorrect), the existing mechanism lacks refined gradient signals for differentiated learning, which directly weakens the distinguishability of learning gradients and ultimately restricts the further improvement of model performance. Secondly, there is the widespread phenomenon of ``difficulty drift" \cite{qu2025can,DBLP:journals/corr/abs-2506-06632,DBLP:journals/corr/abs-2504-05520}that occurs during the training process: with the accumulation of training steps, the model's ability continues to improve, leading to dynamic changes in the relative difficulty of problems\textemdash originally high-difficulty problems gradually transform into medium or low-difficulty ones. In this case, if a fixed weight strategy is still used for undifferentiated training on all problems, it will not only reduce the overall training efficiency but also limit the reasoning upper limit of the model due to insufficient attention given to high-difficulty problems.

To address these challenges, we propose \textbf{CVPO} (\textbf{C}urriculum-guided \textbf{V}alue-\textbf{V}ariance \textbf{P}olicy \textbf{O}ptimization). At the response trajectory level, we find that token-level value-variance in a trajectory correlates with exploration intensity: high variance implies active exploration, low variance indicates rigidity. Furthermore, our theoretical analysis shows this variance bounds the magnitude of policy updates. Based on these findings, we leverage the stochastic nature of trajectory value-variance to optimize the advantage calculation. By incorporating this variance into our advantage function adjustment mechanism, CVPO can dynamically regulate the weight of trajectories according to their reward types. At the question level, we introduce a dynamic curriculum weighting method that adapts to question difficulty. Our method dynamically adjusts the weights of problems based on their difficulty levels. This ensures that the model focuses on tasks matching its current capabilities at each training stage. Our main contributions are as follows:

\begin{itemize}
\item We propose the CVPO method, which constructs a variance-based advantage function adjustment mechanism at the response trajectory level to enhance model exploration. At the problem level, we introduce a dynamic curriculum weighting approach to mitigate the issue difficulty drift during training.

\item We find that the variance of trajectory values influences the model's gradient updates. Our theoretical analysis proves that appropriately leveraging this variance can improve the upper bound of model performance. Empirical results further confirm that this method significantly enhances both the accuracy and robustness of the model.

\item The evaluation results on multiple mathematical benchmarks show that CVPO outperforms other leading methods. Ablation studies further validate the effectiveness of each improvement.
\end{itemize}

\section{Related Works}

Recently, reinforcement learning (RL)-based post-training has become a crucial paradigm for enhancing the reasoning capabilities of large models. Proximal Policy Optimization (PPO)~\cite{schulman2017ppo}, a widely adopted RL algorithm, balances reward maximization and policy stability by incorporating a clipping mechanism and KL regularization to constrain policy updates. However, when dealing with long-chain reasoning tasks, standard PPO faces issues such as value function initialization bias and reward signal decay, limiting its effectiveness in complex reasoning scenarios.

To improve reasoning quality, ORZ \cite{DBLP:journals/corr/abs-2503-24290} incorporates a value model for advantage estimation within the PPO framework but opts for Monte Carlo returns instead of Generalized Advantage Estimation (GAE), sacrificing the bias-variance trade-off provided by GAE, which leads to greater fluctuations during training.

In contrast, DeepSeek-R1 utilizes Group Relative Policy Optimization (GRPO) \cite{2025deepseek-math}, which completely eliminates the value network and uses group-normalized rewards as advantages, achieving efficient training without relying on a value model. GRPO significantly improves training stability and efficiency whilst maintaining performance. However, ``value-free" methods like GRPO cannot leverage trajectory-level value information, which limits their ability to perform fine-grained modeling of the reasoning process.

To address the instability of value function training in long reasoning chains, VC-PPO \cite{DBLP:journals/corr/abs-2503-01491} introduces value pre-training and decoupled GAE computation to mitigate initial value bias and reward decay. Building on this foundation, VAPO \cite{DBLP:journals/corr/abs-2504-05118} further incorporates length-adaptive GAE and token-level loss functions, outperforming value-free methods like GRPO in various mathematical reasoning tasks and thus validating the potential of value networks in complex reasoning tasks.

Despite these advancements, existing value-based approaches still exhibit notable deficiencies in feedback precision. When the generated trajectories within a batch display high homogeneity (e.g., all correct or all incorrect), current algorithms struggle to provide discriminative gradient signals, failing to effectively utilize trajectory-level statistical characteristics (such as value-variance and intra-group differences). The current utilization of value signals has not fully exploited their supervisory potential in structured reasoning tasks. Our work aims to address this limitation by proposing a variance-aware and accuracy-adaptive weighting mechanism to enhance the expressiveness and learning efficiency of value signals.

\section{Preliminaries}
We conceptualize language generation as a token-level Markov Decision Process (MDP) represented by the tuple \(M = (S, \mathcal{A}, \mathbb{P}, R, d_0)\). In this framework, the state space \(S\) consists of states \(s_t = (x_0, \ldots, x_m, y_0, \ldots, y_t)\), which encapsulate all tokens generated up to time step \(t\) — with \(x\) denoting the input prompt and \(y\) representing the generated response. The action space \(\mathcal{A}\) corresponds to a fixed set of discrete vocabulary items, while the dynamics \(\mathbb{P}\) describe deterministic token transitions, specifically \(\mathbb{P}(s_{t+1} | s_t, a) = 1\) when \(a = y_{t+1}\). The reward function \(R\) delivers scalar evaluative feedback, and \(d_0\) represents the initial probability distribution over possible prompts.

Proximal Policy Optimization (PPO) is a popular reinforcement learning algorithm that enhances policy performance through iterative updates while ensuring training stability.

Its core mechanism limits excessive policy changes during updates using a clipped objective function, preventing significant deviations from the previous policy and avoiding the instability issues of other policy gradient methods.

PPO optimizes the clipped surrogate objective:
\begin{equation}
L(\theta) = \mathbb{E}_t\left[ \min(r_t(\theta)\hat{A}_t, \text{clip}(r_t(\theta), 1-\epsilon, 1+\epsilon)\hat{A}_t) \right]
\end{equation}
where \( r_t(\theta) = \frac{\pi_\theta(a_t|s_t)}
{\pi_{\theta_{\text{old}}}(a_t|s_t)} \) represents the ratio of the new policy to the old policy, \( \epsilon \)  controls the clipping range, and \( \hat{A}_t \) is the estimated advantage. The advantage is calculated as:
\begin{equation}
\hat{A}_t = \sum_{k=0}^{T-t-1} (\gamma \lambda)^k \delta_{t+k}
\end{equation}
with \( \delta_t = r_t + \gamma V(s_{t+1}) - V(s_t) \), where \( \gamma \) is the discount factor, \( \lambda \) is the GAE parameter, \( r_t \) is the reward at step \( t \), and \( V(s) \) is the state value function.

\section{Methodology}
\subsection{Stochasticity-Aware Adaptive Advantage Function Correction Mechanism}
\subsubsection{Theoretical analysis of the influence of trajectory value-variance on gradient estimation}

To further understand the mechanism by which CVPO improves advantage estimation through modeling the variance of the value function, we analyze how value-variance along trajectories influences policy updates.

The policy gradient is defined as:
\begin{equation}
\nabla_\theta J(\theta) = \mathbb{E}_{s,a \sim \pi_\theta} \left[ \nabla_\theta \log \pi_\theta(a|s) \cdot A(s,a) \right],
\end{equation}
where the advantage function is given by $A(s,a) = Q(s,a) - V(s)$, with $Q(s,a)$ denoting the state-action value and $V(s)$ the state value.

Assume that the L2 norm of the policy log-probability gradient is bounded by a constant $C$, i.e.,
\begin{equation}
\|\nabla_\theta \log \pi_\theta(a|s)\|_2 \leq C.
\end{equation}

Applying the Cauchy-Schwarz inequality, we derive an upper bound on the gradient norm:
\begin{equation}
\begin{aligned}
\|\nabla_\theta J(\theta)\|_2 
&\leq \mathbb{E}\left[ \|\nabla_\theta \log \pi_\theta(a|s)\|_2 \cdot |A(s,a)| \right] \\
&\leq C \cdot \mathbb{E}\left[ |A(s,a)| \right].
\end{aligned}
\end{equation}

Since $\mathbb{E}[A(s,a)] = 0$, the variance of the advantage function simplifies to:
\begin{equation}
\text{Var}(A) = \mathbb{E}[A(s,a)^2].
\end{equation}

By Jensen’s inequality, we have:
\begin{equation}
\left( \mathbb{E}[|A(s,a)|] \right)^2 \leq \mathbb{E}[A(s,a)^2] = \text{Var}(A),
\end{equation}
which implies:
\begin{equation}
\mathbb{E}[|A(s,a)|] \leq \sqrt{\text{Var}(A)}.
\end{equation}

Expanding the variance of $A(s,a)$:
\begin{equation}
\text{Var}(A) = \text{Var}(Q) + \text{Var}(V) - 2\,\text{Cov}(Q, V).
\end{equation}

Using the Cauchy-Schwarz inequality for covariance,
\begin{equation}
|\text{Cov}(Q, V)| \leq \sqrt{\text{Var}(Q) \cdot \text{Var}(V)},
\end{equation}
we obtain the following upper bound:
\begin{equation}
\begin{aligned}
\text{Var}(A) 
&\leq \text{Var}(Q) + \text{Var}(V) + 2\sqrt{\text{Var}(Q) \cdot \text{Var}(V)} \\
&= \left( \sigma_Q + \sqrt{\text{Var}(V)} \right)^2,
\end{aligned}
\end{equation}
where $\sigma_Q = \sqrt{\text{Var}(Q)}$.

Substituting this into the earlier bound yields:
\begin{equation}
\mathbb{E}[|A(s,a)|] \leq \sigma_Q + \sqrt{\text{Var}(V)}.
\end{equation}

Finally, the magnitude of the policy gradient is bounded by:
\begin{equation}
\|\nabla_\theta J(\theta)\|_2 \leq C \cdot \left( \sigma_Q + \sqrt{\text{Var}(V)} \right).
\end{equation}

Our theoretical analysis establishes a direct link between the magnitude of policy gradients and the variance of the value function. This highlights the critical role of controlling $\text{Var}(V)$ in achieving stable policy optimization, particularly in reinforcement learning and large language model reasoning tasks. We show that high value variance leads to larger gradients. When this variance is modulated with respect to trajectory quality, selectively increasing it can amplify learning signals in the right direction. This insight provides a principled approach to enhancing both learning efficiency and convergence stability.

\subsubsection{Analysis: Value Stochasticity in LLM Reasoning }

In the field of reinforcement learning for large models, balancing exploration and exploitation remains a key bottleneck that restricts the efficiency of policy optimization. Existing advantage function designs have two significant flaws: 

\begin{enumerate}
    \item They lack a dynamic response mechanism to the intrinsic stochasticity of generated trajectories, and the fixed-weight scheme struggles to meet the differentiated stochasticity demands of trajectories with varying quality (correct/incorrect responses). 
    
    \item They fail to incorporate statistical characteristics of value estimation (such as variance) into the policy optimization framework, resulting in positive samples being difficult to converge stably due to excessive exploration, and negative samples falling into local optima due to rigid penalty mechanisms, with limited exploration depth.
\end{enumerate}

To address the above issues, we propose a stochasticity-aware adaptive advantage function correction mechanism. It quantifies the intrinsic stochasticity of the generation process through the variance of trajectory value estimates, and achieves precise regulation of ``strong constraints on positive samples and promotion of exploration on negative samples" by designing a dynamic weight function ($W_{\text{S}}$) to differentially weight token-level advantage values.

\begin{wrapfigure}{r}{0.5\linewidth}
\centering
\includegraphics[width=\linewidth]{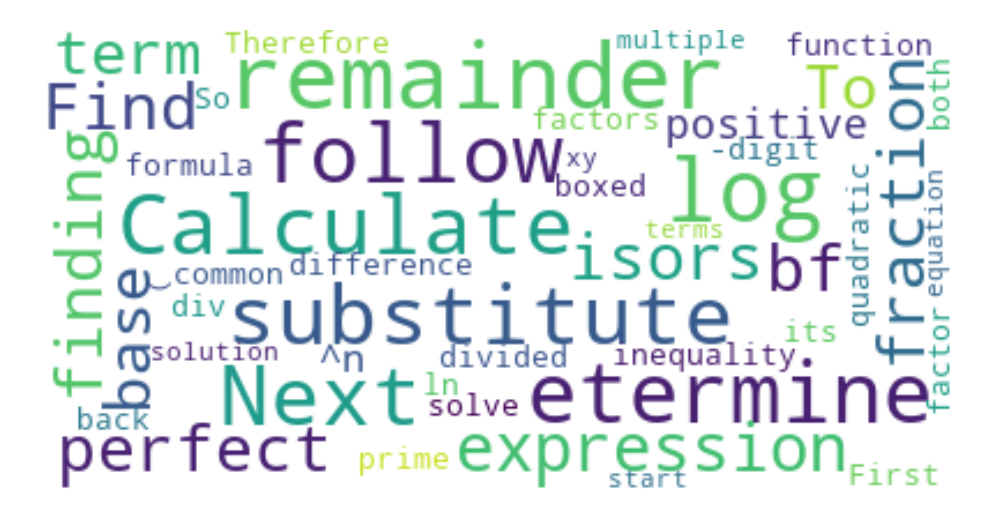}
\caption{Tokens with the highest average values from generated responses in the training set. They are closely related to the training task, which is mathematics in our experiment setting}
\label{fig:token-word-cloud}
\end{wrapfigure}

As shown in Figure \ref{fig:token-word-cloud}, the training task strongly influences the state values of the generated tokens and the resulting states they create, with mathematics terminology obtaining high values for our experiments dealing with math subjects. Inversely, low-value tokens tend to be less relevant to the training task. By monitoring changes in token-level values through the variance, we can use it to measure the amount of exploration being done within a single response trajectory. High variance indicates greater stochasticity and exploration, whereas low variance means limited exploration and more rigid responses.

\subsubsection{Method: Stochasticity-Aware Advantage Correction}

The variance of trajectory ($\tau$) value estimates ($\text{Var}(\tau)$) is used as the stochasticity quantification index, and an asymmetric weight adjustment mechanism is constructed for trajectories of different reward types ($R(\tau)=1$ for positive samples, $R(\tau)=0$ for negative samples):

Let $\mathcal{B} = \{(v_i, r_i, \text{idx}_i)\}_{i=1}^B$ denote a batch of $B$ trajectories, where $v_i \in \mathbb{R}^{T_i}$ is the sequence of estimated state values for the $i$-th sample, $r_i \in \mathbb{R}^{T_i}$ is the corresponding token-level reward, and $\text{idx}_i \in \mathbb{N}^+$ denotes the index of the associated prompting template or question instance. We define the value-variance signal for each sample as:
$$
\sigma_i = \mathrm{std}(v_i),
$$
which measures the intra-trajectory uncertainty of the value estimation.

To normalize this signal across samples from the same prompt, we group samples by their index. Let $\mathcal{G}_k = \{ i \mid \text{idx}_i = k \}$ denote the set of sample indices corresponding to prompt $k$. The mean value standard deviation for prompt $k$ is then computed as:
\begin{equation}
\mu_k = \displaystyle \frac{1}{|\mathcal{G}_k|} \sum_{i \in \mathcal{G}_k} \sigma_i
\label{eq:group_mean}
\end{equation}

This group-wise mean $\mu_k$ serves as a reference point to assess whether a given sample exhibits higher or lower value fluctuation compared to others under the same prompting context.

For trajectories with correct responses, high stochasticity needs to be suppressed to accelerate convergence. When $\text{Var}(\tau)$ exceeds the threshold ($\mu_P$), the weight function enhances the penalty; otherwise, rewards are given to reinforce low-stochasticity strategies. The function is defined as:

\begin{footnotesize}
\begin{equation}
W_{\text{S}}(\text{Var}(\tau), 1) = 1.0 + \alpha_P \cdot \text{sigmoid}(\lambda_P (\text{Var}(\tau) - \mu_k))
\end{equation}
\end{footnotesize}
where $\lambda_P$ is a fixed steepness parameter, controlling the sensitivity of the weight to changes in variance; $\alpha_P$ is a fixed gain parameter, adjusting the dynamic range of reward and penalty intensities.

For trajectories with incorrect responses, high stochasticity needs to be retained to expand the exploration space. When $\text{Var}(\tau)$ is below the threshold ($\mu_N$), the weight function increases the penalty intensity; otherwise, the penalty is reduced to encourage high-stochasticity exploration. The function is defined as:

\begin{footnotesize}
\begin{equation}
W_{\text{S}}(\text{Var}(\tau), 0) = 1.0 - \alpha_N \cdot \text{sigmoid}(\lambda_N (\mu_k - \text{Var}(\tau)))
\end{equation}
\end{footnotesize}
where $\lambda_N$ is a fixed steepness parameter, regulating the penalty sensitivity of the weight to low-variance trajectories; $\alpha_N$ is a fixed penalty amplitude parameter, limiting the adjustment range of penalty intensity for negative samples.

\subsection{Difficulty-guided Advantage Weighting}
\subsubsection{Analysis: Difficulty Shift in Reinforcement Learning for Large Language Models}
In reinforcement learning-based training, models tend to prioritize solving easy problems—due to their readily obtainable reward signals—leading to insufficient reasoning length allocated to complex problems, and sometimes even bypassing reasoning steps entirely by directly outputting answers. However, as training progresses, the model's capabilities gradually improve, and problems initially perceived as ``hard'' may evolve into ``medium'' or ``easy'' ones. Training all difficulty levels with fixed weights results in overfitting to simple samples in later stages, reducing training efficiency. Moreover, the model's insufficient focus on difficult problems limits the potential for further improvement in its overall performance.

Additionally, existing curriculum learning (CL) methods typically rely on predefined, static difficulty metrics, which fail to account for the dynamic shift in the model's perception of task difficulty during training—referred to as the ``difficulty shift'' phenomenon. This misalignment between the predefined curriculum and the model's real-time learning needs negatively impacts both learning efficiency and final performance.

To address these challenges, we propose a \textbf{difficulty-aware dynamic curriculum weighting} method. Our approach continuously assesses sample difficulty during training and dynamically adjusts the weights of problems with different difficulty levels in the advantage computation, enabling the model to focus on tasks that match its current capability at each training stage. This effectively prevents the model from falling into the ``easy-problem trap'' and promotes sustained learning on complex reasoning tasks.

\subsubsection{Method: Difficulty-Aware Dynamic Curriculum Weighting}

First, we introduce a Bayesian inference framework to maintain a posterior distribution over the accuracy for each question, and further aggregate these to obtain a dynamic estimate of the model's current capability. During the model training process, we employ a Bayesian inference framework to dynamically track the success rate of each sample $\tau$. Specifically, we set a Beta prior distribution for the initial success rate $\gamma_{\tau}^{0}$ of each sample as
\begin{equation}
    \gamma_{\tau}^{0} \sim \text{Beta}(\alpha_{\tau}^{0}, \beta_{\tau}^{0}),
\end{equation}
where $\alpha_{\tau}^{0} = \beta_{\tau}^{0} = 1$, which corresponds to a uniform prior.

After the model generates responses to these samples, we calculate the number of successes:
\begin{equation}
    s_{\tau}^{0} = \sum_{j=1}^{k} r_{\tau}^{0,j},
\end{equation}
where $r_{\tau}^{0,j} \in \{0, 1\}$ are binary rewards (1 for correct, 0 for incorrect). Using this, we obtain the initial posterior distribution via a recursive update rule:
\begin{equation}
    \gamma_{\tau}^{0} \mid H_0 \sim \text{Beta}(\alpha_{\tau}^{0'}, \beta_{\tau}^{0'}),
\end{equation}
with updated parameters
\begin{align}
    \alpha_{\tau}^{0'} &= \alpha_{\tau}^{0} + s_{\tau}^{0}, \\
    \beta_{\tau}^{0'} &= \beta_{\tau}^{0} + k - s_{\tau}^{0},
\end{align}
where $H_0$ denotes the initial optimization history.

To monitor the model's learning dynamics, we define a trend indicator $\Delta_{\tau}^{t}$ as:
\begin{equation}
    \Delta_{\tau}^{t} = \frac{\bar{\gamma}_{\tau}^{t} - \bar{\gamma}_{\tau}^{t-1}}{\bar{\gamma}_{\tau}^{t-1}},
\end{equation}
where $\bar{\gamma}_{\tau}^{t} = \frac{\alpha_{\tau}^{t'}}{\alpha_{\tau}^{t'} + \beta_{\tau}^{t'}}$ is the posterior mean of sample $\tau$ at training round.

Based on this indicator, we define two stagnation conditions to detect a learning bottleneck:

For the low-accuracy sample set
\begin{equation}
    T_{\text{low}} = \left\{ \tau \mid \bar{\gamma}_{\tau}^{t} \leq 0.3 \right\},
\end{equation}

If samples with accuracy below 30\% satisfy $|\Delta_{\tau}^{t}| \leq 0.05$, the model is considered to be in a stagnant state in the low-accuracy region.

For the high-accuracy sample set
\begin{equation}
    T_{\text{high}} = \left\{ \tau \mid \bar{\gamma}_{\tau}^{t} \geq 0.8 \right\},
\end{equation}

If samples with accuracy above 80\% satisfy $\Delta_{\tau}^{t} \leq 0.03$, the model is considered to be experiencing a slowdown in accuracy improvement at a high-performance level.

When both conditions are simultaneously satisfied, the model is judged to have entered a learning bottleneck, and a shift in the learning strategy is triggered.

We introduce the function $ W_{\textit{D}}(\cdot)$ to adapt to the trained policy's mastery of currently sampled question, allowing the policy to learn gradually from training data and focus on the most difficult problems with regards to its abilities.

In the initial stage of training, to promote rapid learning and effective performance improvement, we place special emphasis on tasks that the model is capable of solving but currently performs poorly on. By focusing more attention on questions with an accuracy rate around 0.5, we leverage their high reward variance and relatively high lower bound of KL divergence, thereby obtaining stronger gradient signals. This strategy not only accelerates the optimization of the policy but also ensures that the model rapidly acquires effective knowledge and skills during the early stages of training.

$ W_{\textit{D}}(\cdot)$ is defined as follows:

\begin{equation}
W_{\textit{D}}(\rho_q) = \frac{2}{1 + e^{-k_1 (\rho_q - c_1)}} \cdot \frac{1}{1 + e^{k_2 (\rho_q - c_2)}},
\end{equation}

where hyperparameters$ k_1 = 1.6 $, $ c_1 = -0.2 $, $ k_2 = 2 $, and $ c_2 = 1.2 $, and $\rho_q$ is the most recent accuracy for sampled question $q$.

When the model enters a learning bottleneck, it has typically already mastered simple problems but exhibits insufficient exploration of difficult ones. To guide the model toward breaking through this plateau, we dynamically shift the difficulty weighting distribution to the left---that is, toward questions with lower accuracy---thereby increasing the learning weight on high-difficulty, low-accuracy problems. Meanwhile, to prevent the model from falling into the ``easy-problem trap,'' we set the weights of questions with accuracy $1$ to zero, effectively suppressing redundant training on fully mastered samples and encouraging the model to focus its attention on challenging, unsolved tasks.

At this point, the difficulty weighting function $W_{\text{D}}(\cdot)$ is defined as follows:

\begin{equation}
W_{\text{D}}(\rho_q) = \frac{2\left(1 - \rho_q^2\right)}{1 + e^{-k'_1 (\rho_q - c'_1)}} \cdot \frac{1}{1 + e^{k'_2 (\rho_q - c'_2)}},
\end{equation}

where the hyperparameters are $k'_1 = 4.1$, $c'_1 = -0.1$, $k'_2 = 2$, $c'_2 = 0.9$, and $\rho_q$ is the most recent accuracy for sampled question $q$.

\subsection{Unified Advantage Function with Stochasticity and Difficulty Awareness}

Based on the proposed methods of trajectory variance and dynamic curriculum learning, we derive the final weighted advantage function as follows:

\begin{equation}
A_{t'}' = W_{\text{S}}(\text{Var}(\tau), R(\tau)) \cdot W_{\text{D}}(\rho_q, t_c) \cdot A,
\label{eq:weighted_advantage_english}
\end{equation}

where $\tau$ is the output trajectory for sampled question $q$, $t_c$ is the current phase in curriculum training, $\text{Var}(\tau)$ is the variance for the values of trajectory $\tau$, and $R(\tau)$ is the overall return of the trajectory.

\section{Experiment}
\subsection{Experiment Setup}

\textbf{Datasets and Settings.} We use DAPO-Math-17k~\citep{2025dapo-rl} as the training set and evaluate on datasets including AIME 2024, AIME 2025, AMC 2023, MATH-500 \citep{hendrycksmath500}, and AMC 2024. For the AMC datasets, we filter out questions that require comprehending figures to answer. Performance is evaluated as the mean accuracy over 32 attempts, \textit{avg@32}.

\textbf{Reward Setting.} We adopt a rule-based reward mechanism to provide a stable and explicit training signal for the reinforcement learning-based post-training process. 
Specifically, the model's output is required to follow a predefined structured format: the final answer must be placed within the \texttt{\textbackslash boxed\{\}} tag or after the text \texttt{Answer:} on a separate new line. 
The reward signal is determined by the correctness of the final answer: if the predicted answer exactly matches the ground truth, the model receives a reward of $+1$; otherwise, it receives $-1$.

\begin{table}[tb]
    \centering
    \begin{tabular}{>{\RaggedRight\arraybackslash}p{34ex} 
    >{\Centering\arraybackslash}p{8ex}
    >{\Centering\arraybackslash}p{8ex}
    >{\Centering\arraybackslash}p{8ex}
    >{\Centering\arraybackslash}p{8ex}
    >{\Centering\arraybackslash}p{12ex}}
    \toprule    
    Methods & AIME24 & AIME25 & AMC23 & AMC24 & MATH500 \\
    \midrule
    Qwen2.5-7B & 0.046 & 0.011 & 0.198 & 0.201 &0.251 \\
    GRPO (ASYN) & 0.137 & 0.102 & 0.468 & 0.383 & 0.665 \\
    VAPO & 0.144 & 0.046 & 0.441 & 0.260 & 0.452 \\
    CVPO (only Stochasticity) & \underline{0.188} & \underline{0.100} & \underline{0.676} & \underline{0.497} & \underline{0.765} \\
    CVPO (only Dynamic Curriculum) & 0.162& 0.065 & 0.538 & 0.412 & 0.558 \\
    CVPO (w/ 0.3 variance coeff) & \textbf{0.237} & \textbf{0.133} & \textbf{0.720} & \textbf{0.512} & \textbf{0.802} \\
    \bottomrule
    \end{tabular}
    \caption{\textit{avg@32} accuracy for algorithms across math benchmarks. Best results are in \textbf{bold}, second best are \underline{underlined}}
    \label{tab:benchmark_res}
\end{table}

\textbf{Implementation Details.} We enhance the mathematical reasoning capability of the Qwen2.5-7B model using the CVPO algorithm, and include VAPO and GRPO (ASYN) for comparison purposes. 
All experiments are conducted on 8 server nodes, each equipped with 8 NVIDIA A800 GPUs. 
We use AdamW as the optimizer, with the actor learning rate set to $1 \times 10^{-6}$ and the critic learning rate to $2 \times 10^{-6}$. 
The training batch size is 256, with each rollout batch containing 16 responses, and the maximum response length is 6144 tokens. 
We forego the use of KL constraints in our setup as we found this to be limiting model updates when the trained policy generates long reasoning traces.
We train our value model from Qwen2.5-7B, utilizing a checkpoint that has a value-model loss less than 1 and an explained variance greater than 0.
We use a warm-up schedule of 21 steps for the value model checkpoint stated previously, set the GAE $\lambda_{critic} = 0.1$, the length-adaptive GAE factor $\alpha = 0.17$, the discount factor $\gamma$ to 1.0, clip coefficients to $\epsilon_{lower} = 0.2$ and $\epsilon_{higher} = 0.28$, and positive LM loss coefficient $\mu = 0.1$.
For comparison purposes with value-free algorithms, we also included GRPO in our experiments. However, in our attempts, we experienced extreme response-length collapse using the generic GRPO algorithm. Therefore, we utilized a modified version of GRPO\textemdash single-step async GRPO (ASYN) \textemdash to ensure representative comparisons with our value-based approaches. Hyperparameters such as training batch size and amount of rollout remain unchanged, and an uniform clipping coefficient of $\epsilon=0.2$ is used.

\begin{figure}[htbp]
    \centering
    \includegraphics[width=0.72\textwidth]{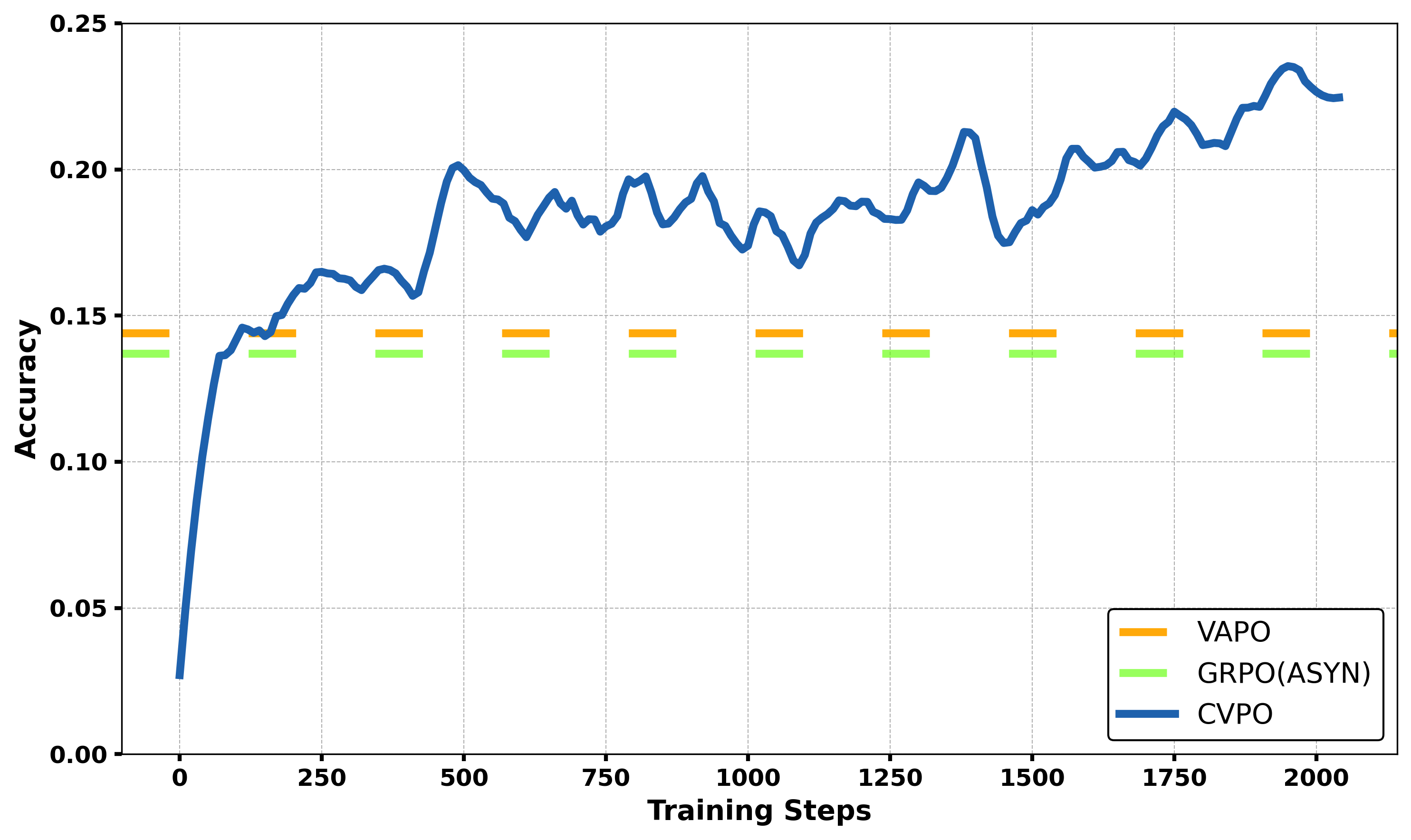} 
    \caption{Comparison of CVPO and previous SOTA methods on the AIME 2024 benchmark using the Qwen2.5-7B model. The x-axis represents training steps. CVPO consistently demonstrates a significant advantage throughout the training process.}
    \label{fig:train}
\end{figure}

\section{Results and Discussions}

\subsection{Results}

As shown in Table~\ref{tab:benchmark_res}, our proposed method \textbf{CVPO} significantly outperforms existing SOTA methods across multiple datasets. Specifically, it achieves improvements of \textbf{9.3\%}, \textbf{3.1\%}, \textbf{25.2\%}, \textbf{12.9\%}, and \textbf{13.7\%} over the second-best methods on AIME24, AIME25, AMC23, AMC24, and MATH500, respectively. These results demonstrate the strong effectiveness of CVPO in complex mathematical reasoning tasks.

We also conduct ablation studies on the two core components of CVPO. The variant with only stochasticity-aware advantage correction (\textit{CVPO (only Stochasticity)}) and the variant with only difficulty-aware dynamic curriculum weighting (\textit{CVPO (only Dynamic Curriculum)}) both outperform the baseline method \textit{VAPO}. This confirms the effectiveness of each individual component. Among them, the stochasticity-aware correction brings greater performance gain. This is because it expands the upper bound of model exploration.

As shown in Figure~\ref{fig:train}, CVPO, VAPO, and GRPO (ASYN) are all trained for over 2000 steps. CVPO focuses on ``medium-difficulty'' problems in the early stage and uses trajectory variance to adjust advantage estimation, which helps boost performance quickly. In the mid-training stage, when performance plateaus, CVPO changes the advantage function to focus more on hard problems. This improves the model’s reasoning ability. As a result, CVPO continues to improve in accuracy after 1500 training steps, showing its more effective learning strategy.

\subsection{Discussions}

\begin{figure}[htbp]
\begin{center}
\includegraphics[width=0.72\textwidth]{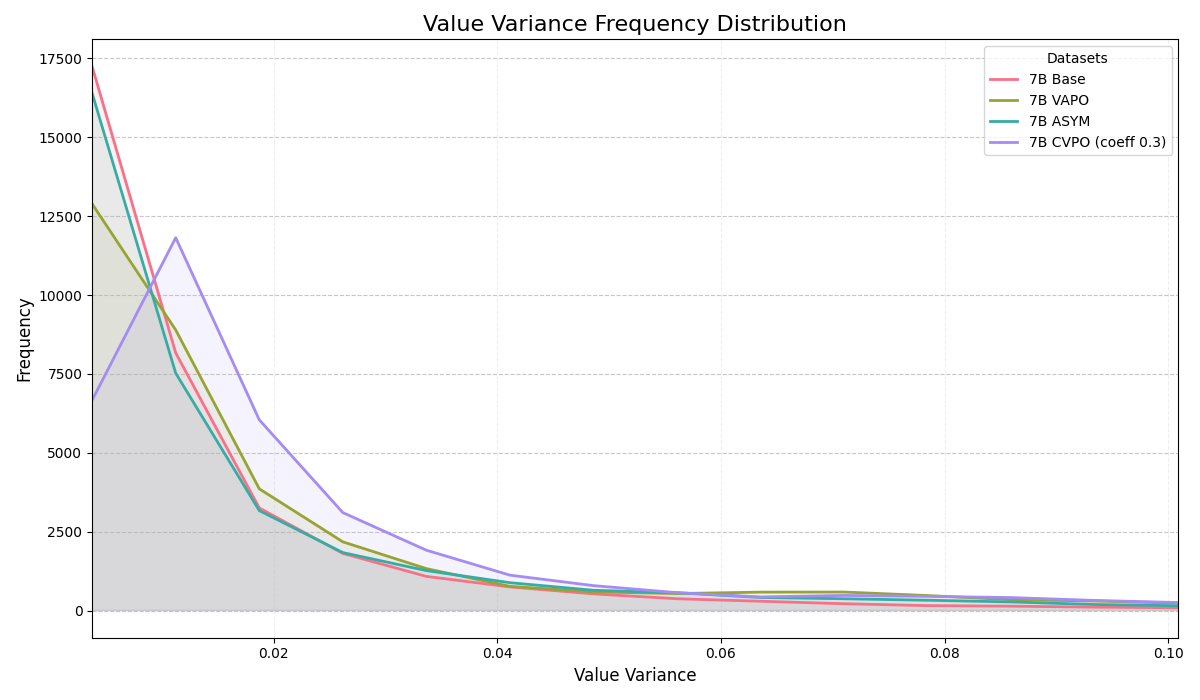}
\end{center}
\caption{Trajectory-level value-variance distribution on the training set between Qwen2.5 7B Base and evaluated algorithms. The distribution steadily shifts towards higher values from the base model to value-based algorithms}
\label{fig:val-var-comparison}
\end{figure}

Figure \ref{fig:val-var-comparison} shows the distribution of trajectory-level value-variance on the training set.
Predictably, the base Qwen2.5-7B model displays the least value-variance on average, reflecting the lack of exploration within response trajectories. 
Training via GRPO-ASYN resulted in incrementally more responses with higher value-variances, but due to the lack of a value model to guides policy updates on a finely-grained level, the difference is not too pronounced.
Coming to the value-based VAPO method, we can observe a noticeable shift in value-variance distribution towards higher values. This shift is much more visible with CVPO, 
This shift in value-variance distribution towards higher values indicates a greater degree of exploration in the generated response trajectories, which can potentially assist models to generate correct answers by considering more possible pathways.

\begin{figure}[hb]
\begin{center}
\includegraphics[width=0.72\textwidth]{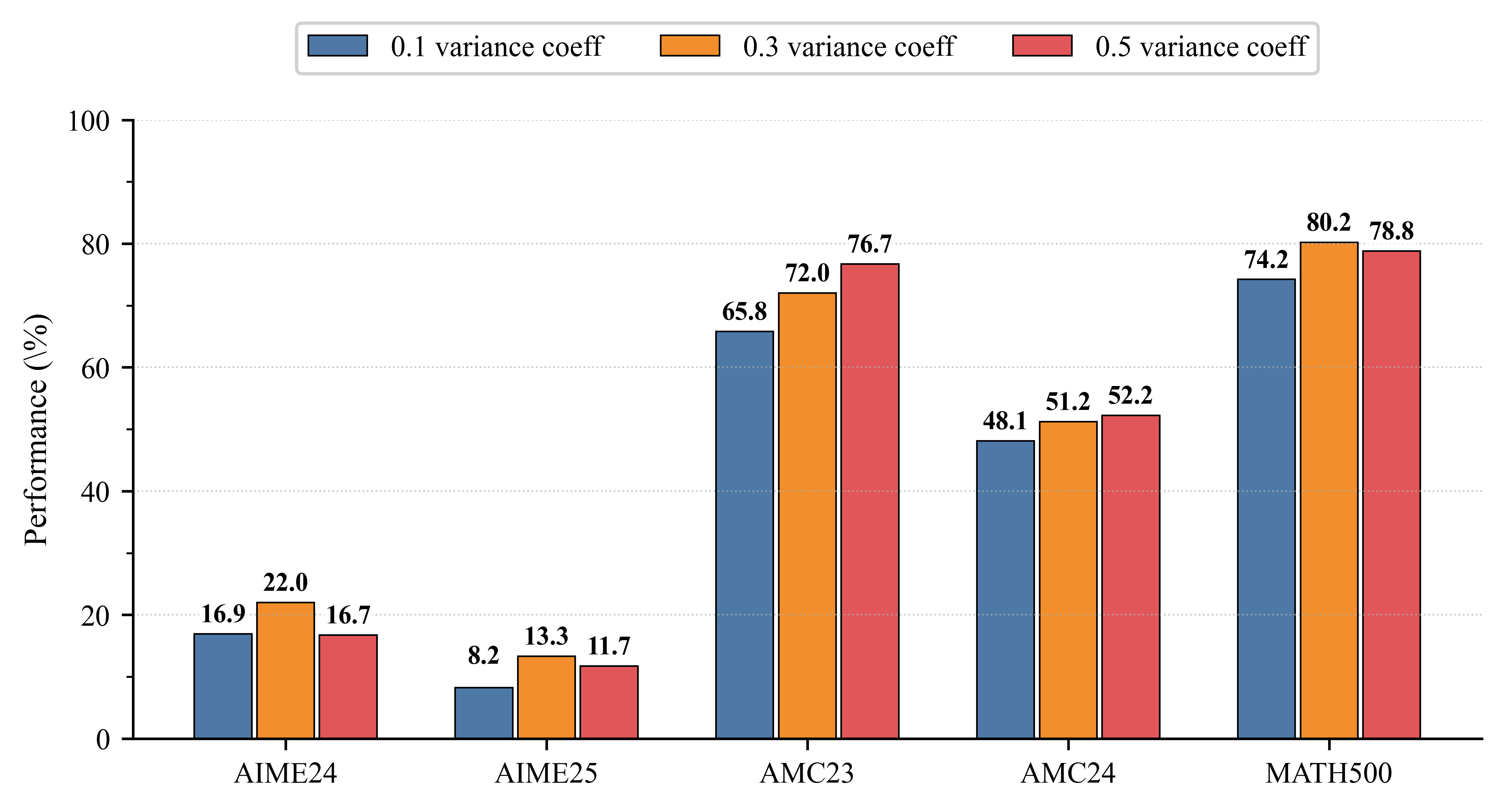}
\end{center}
\caption{CVPO performance using different value-variance coefficients. Higher coefficient values display better performance for AMC tests, but less performance for harder AIME questions}
\label{fig:cvpo-coeff-comp}
\end{figure}

To investigate the impact of stochasticity-aware adaptive advantage weighting, we conduct ablation studies by evaluating performance using different value-variance coefficients. As shown in Figure~\ref{fig:cvpo-coeff-comp}, higher coefficients values generally led to better performance, but there exists a crossover point where increasing coefficients actually results in degraded performance. This falls in line with our expectation, as the coefficients directly affect the magnitude of policy loss. With a bigger value and thereby bigger policy loss, the policy model undergoes larger updates which impacts convergence and undermines the model's ability to adapt tasks that require complex and elaborate reasoning processes. This can be seen in the results where performance on AMC increased steadily along with value-variance coefficients but deteriorated on the harder questions of AIME. 

\begin{figure}[htbp]
\centering
\begin{subfigure}[t]{0.49\textwidth}
  \centering
  \includegraphics[width=\linewidth, height=4.0cm, keepaspectratio]{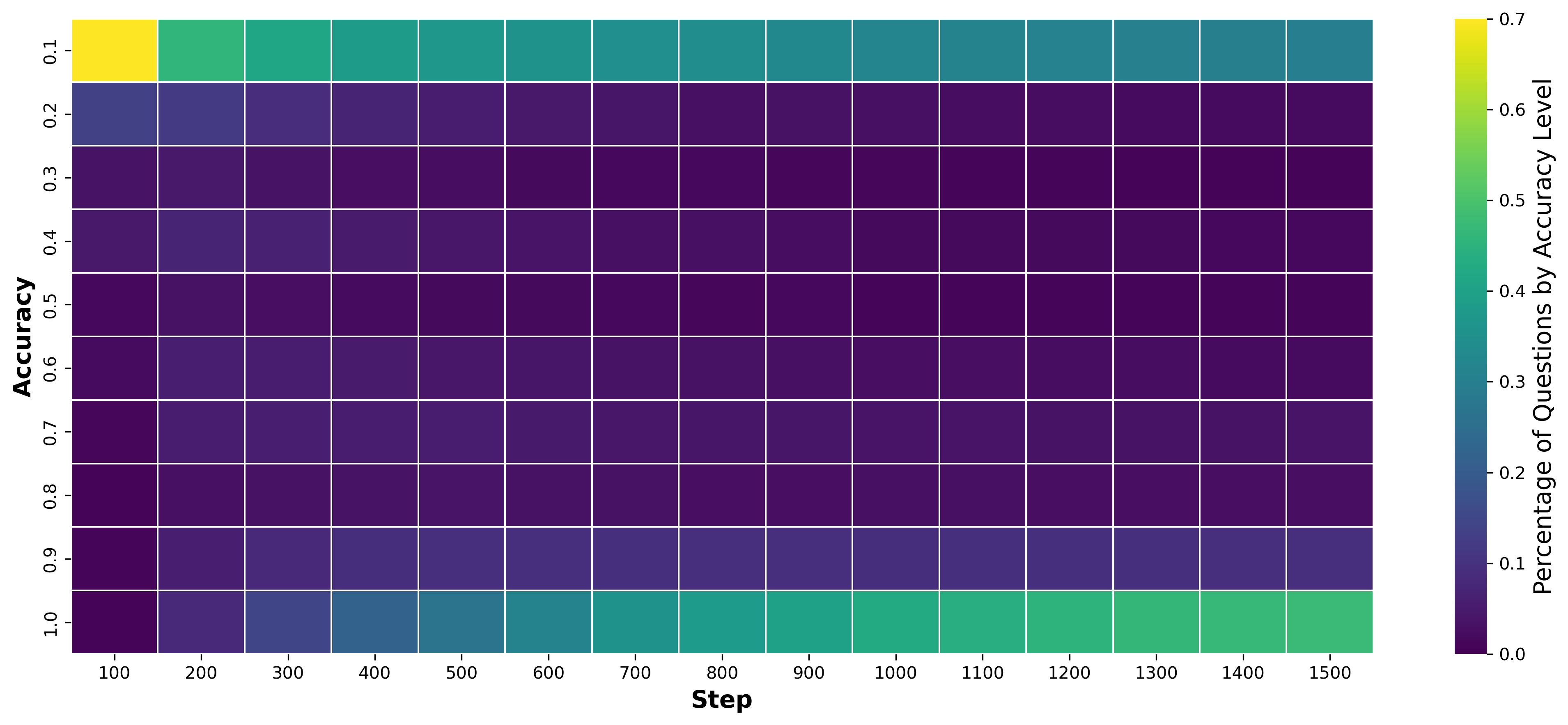}
  \caption{Without Dynamic Question Difficulty Weighting}
  \label{fig:sub1}
\end{subfigure}
\hfill
\begin{subfigure}[t]{0.49\textwidth}
  \centering
  \includegraphics[width=\linewidth, height=4.0cm, keepaspectratio]{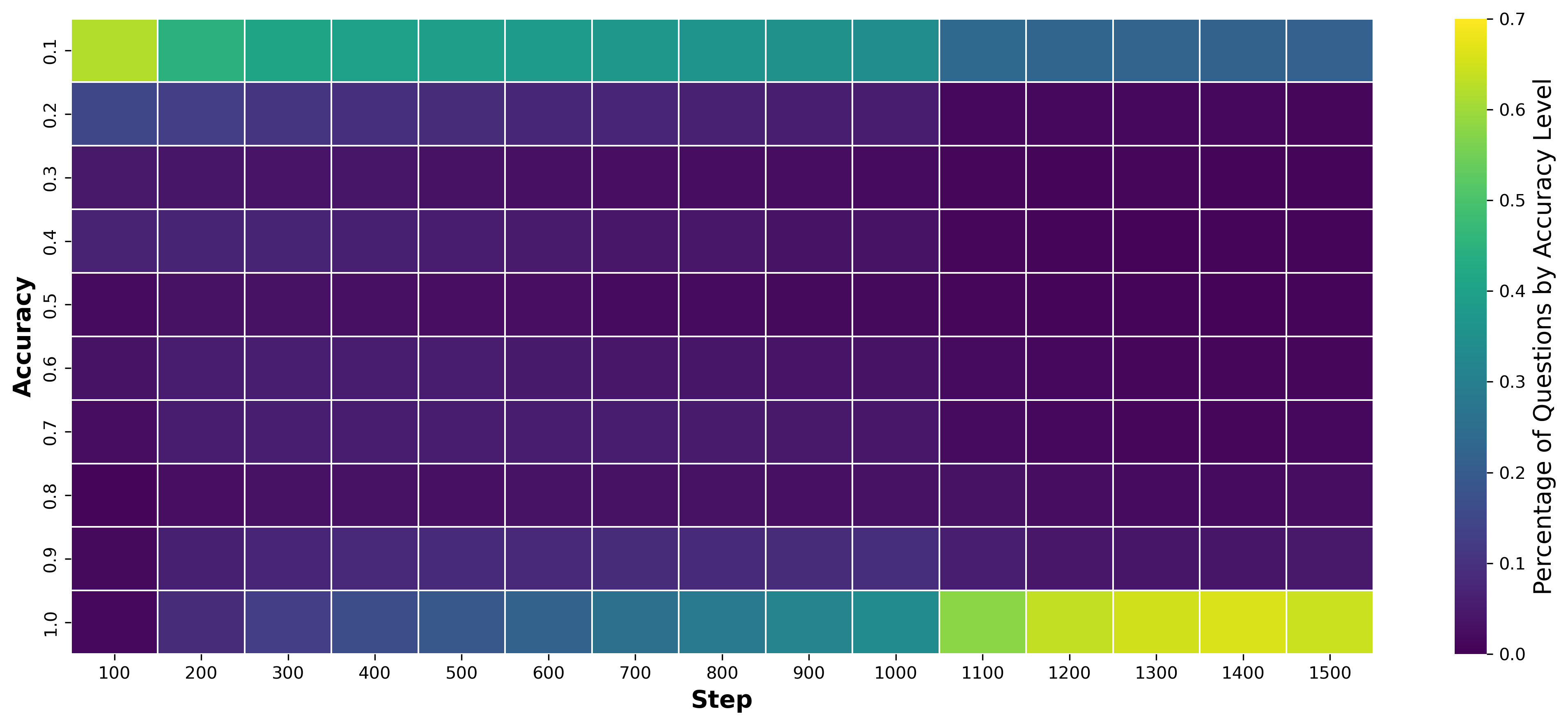}
  \caption{With Dynamic Question Difficulty Weighting}
  \label{fig:sub2}
\end{subfigure}
\caption{Comparison of training with and without dynamic question difficulty weighting. The color in each cell represents the proportion of questions within that accuracy bin relative to the total number of questions. Warmer colors (yellow) indicate a higher proportion, while cooler colors (blue) indicate a lower proportion.}
\label{fig:heatmap-comparison}
\end{figure}

As shown in Figure~\ref{fig:heatmap-comparison} (a), without the difficulty-aware dynamic curriculum weighting mechanism, the model tends to fall into a "simple-question trap". It mainly focuses on easier questions and pays insufficient attention to harder ones. This limits the model's exposure to challenging tasks and constrains its overall performance ceiling. In contrast, as illustrated in Figure \ref{fig:heatmap-comparison} (b), the model focuses more on questions with accuracy around 50\% during early training. This promotes rapid knowledge convergence. When a training bottleneck occurs (e.g., around step 1000), the model dynamically adjusts its attention to different difficulty levels. It increases updates for difficult questions and reduces updates for simple ones. 
As shown in the Figure \ref{fig:heatmap-comparison} (b), this adaptive adjustment enables the model to achieve a higher performance ceiling.

\section{Conclusion}
In this paper, we address two critical challenges faced by existing value-based reinforcement learning methods for enhancing the reasoning capabilities of large language models: insufficient precision in feedback on generated answer trajectories and the phenomenon of problem difficulty drift. To tackle these issues, we propose CVPO (Curriculum-guided Value-Variance Policy Optimization), which incorporates a variance-based advantage function adjustment mechanism at the response trajectory level and a dynamic curriculum weighting method at the problem level.

At the trajectory level, we observe that the token-level variance of values along a trajectory is correlated with the degree of exploration. Our theoretical analysis further proves that the variance of trajectory values affects the model’s policy gradient updates. Based on this, we design an adaptive advantage function that uses the variance of trajectory values to perform differential weighting across trajectories. This function adjusts the weights according to the type of reward, encouraging exploration of incorrect responses while stabilizing convergence on correct ones. Both theoretical analysis and empirical results show that this method significantly improves model accuracy and robustness.

At the question level, CVPO uses a dynamic curriculum weighting scheme. This scheme adapts to the difficulty of each question. During training, it continuously assesses the difficulty of samples. It then adjusts the weights of problems with different difficulty levels. This ensures the model focuses on tasks that match its current ability at each stage. As a result, it reduces difficulty drift and improves training efficiency.

Our experiments on multiple mathematical benchmarks, including AIME 2024, AIME 2025, AMC 2023, MATH-500, and AMC 2024, demonstrate that CVPO achieves higher performance and greater exploration capabilities, leading to more accurate and robust reasoning in language models across various mathematical tasks.

\bibliographystyle{acl_natbib}
\bibliography{custom}

\end{document}